\documentclass[conference]{IEEEtran}
\IEEEoverridecommandlockouts
\usepackage{cite}
\usepackage{amsmath,amssymb,amsfonts}
\usepackage{algorithmic}
\usepackage{graphicx}
\usepackage{textcomp}
\usepackage{xcolor}
\usepackage{times}
\usepackage{epsfig}
\usepackage{amsmath}
\usepackage{amssymb}

\usepackage{diagbox}
\usepackage{multirow}
\usepackage{subfig}
\usepackage{bbm}
\usepackage{float}
\def\BibTeX{{\rm B\kern-.05em{\sc i\kern-.025em b}\kern-.08em
    T\kern-.1667em\lower.7ex\hbox{E}\kern-.125emX}}
\begin{document}

\title{Attention Guided Cosine Margin For Overcoming Class-Imbalance in Few-Shot Road Object Detection\\
}

\author{\IEEEauthorblockN{Ashutosh Agarwal* \thanks{* Work done as an intern at Intel.}} 
\IEEEauthorblockA{ \textit{IIT Delhi}\\
ashutosh.agarwal@cse.iitd.ac.in}
\and
\IEEEauthorblockN{Anay Majee \qquad
Anbumani Subramanian}
\IEEEauthorblockA{ \textit{Intel Corporation}\\
firstname.lastname@intel.com}
\and
\IEEEauthorblockN{Chetan Arora}
\IEEEauthorblockA{ \textit{IIT Delhi}\\
chetan@cse.iitd.ac.in}
}

\maketitle

\begin{abstract}
    Few-shot object detection (FSOD) localizes and classifies objects in an image given only a few data samples.
Recent trends in FSOD research show the adoption of metric and meta-learning techniques, which are prone to catastrophic forgetting and class confusion. 
To overcome these pitfalls in metric learning based FSOD techniques, we introduce \textbf{A}ttention \textbf{G}uided \textbf{C}osine \textbf{M}argin (\textbf{AGCM}) that facilitates the creation of tighter and well separated class-specific feature clusters in the classification head of the object detector.
Our novel \textbf{A}ttentive \textbf{P}roposal \textbf{F}usion (\textbf{APF}) module minimizes catastrophic forgetting by reducing the intra-class variance among co-occurring classes. At the same time, the proposed Cosine Margin Cross-Entropy loss increases the angular margin between confusing classes to overcome the challenge of class confusion between already learned (base) and newly added (novel) classes.
We conduct our experiments on the challenging India Driving Dataset (IDD), which presents a real-world class-imbalanced setting alongside popular FSOD benchmark PASCAL-VOC.
Our method outperforms State-of-the-Art (SoTA) approaches by up to 6.4 $mAP$ points on the IDD-OS and up to 2.0 $mAP$ points on the IDD-10 splits for the 10-shot setting. 
On the PASCAL-VOC dataset, we outperform existing SoTA approaches by up to 4.9 $mAP$ points.

\end{abstract}

\section{Introduction}
    \label{intro}
    Deep Convolution Neural networks (ConvNets) trained on large-scale image datasets \cite{imagenet, coco}, have shown exemplary performance on tasks like classification and object detection \cite{fast-rcnn, yolo2, faster-rcnn}. 
A noticeable pitfall in ConvNets is the requirement of large-scale annotated datasets to achieve State-of-The-Art (SoTA) performance which is both expensive and labor-intensive to acquire. 

\begin{figure}[h]
    \centering
    \includegraphics[width=0.70\columnwidth]{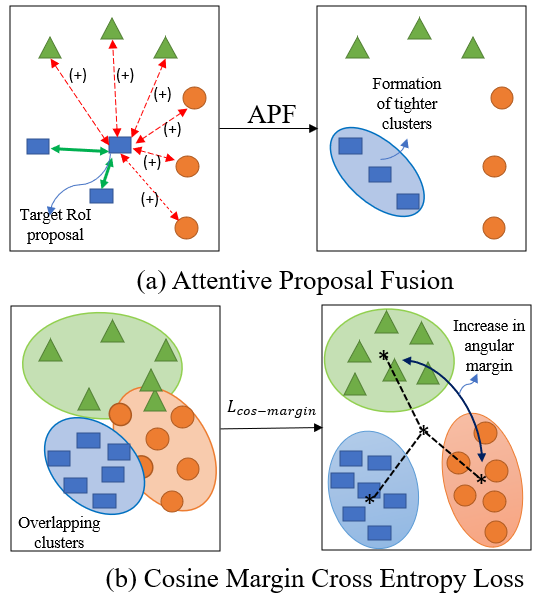}
    \caption{\small Overview of our proposed Attention Guided Cosine Margin approach (a) Visually similar classes are driven closer using our Attentive Proposal Fusion Module (APF). (+) represents that the distance between corresponding RoI proposals should be increased.  (b) Our cosine margin cross entroy loss increases the inter-class margin among classes.}
    \label{fig:intro}
\end{figure}

Recent developments in Machine Learning research have shown significant progress in few-shot learning, especially for image recognition \cite{maml, nonforget, fomaml, fsgraph, protonet, relation-net, matching-net} tasks where algorithms learn to recognize images from limited (few-shot) data samples. On the contrary, \textbf{F}ew-\textbf{S}hot \textbf{O}bject \textbf{D}etection (\textbf{FSOD}) emerges as a relatively unexplored and complex field as it encompasses both localization and recognition tasks. 

Early attempts in FSOD have been made by drawing inspiration from two primary learning strategies in image classification - Meta-Learning \cite{reweight, cme, metarcnn, addfeat} and Metric Learning \cite{fscontrastive, fsdet, pnpdet}. Benchmark experiments conducted by these works show that metric learners significantly outperform meta-learners \cite{metadet} in adapting to few-shot data.
However, the success of metric learners is seldom overshadowed by two dominant issues - \textit{class confusion} and \textit{Catastrophic forgetting}. 
Class confusion refers to misclassifying a predicted Region of Interest (RoI) as an incorrect class label. It is commonly observed among objects belonging to the newly added (novel) classes, which are classified as one or more already learned (base) classes.     
Catastrophic forgetting refers to the degradation in performance of the base classes while adapting to novel classes.
The issues mentioned above become more evident in real-world scenarios such as autonomous driving \cite{city, idd} where only a few data samples are available for detecting less-occurring road objects with significant variations in structure and orientations.

Through extensive experimentation on several SoTA FSOD methods, we observe significant overlaps between feature representations of the base and novel classes. This overlap can be attributed to increasing class confusion in FSOD.
On the other hand, catastrophic forgetting among base classes is a result of FSOD techniques \cite{fscontrastive, fsdet} overfitting to few-shot data samples.

We propose a metric-learning based \textbf{A}ttention \textbf{G}uided \textbf{C}osine \textbf{M}argin (\textbf{AGCM}) approach that exploits the overlapping features among RoI proposals in FSOD to create compact and well-separated class-specific clusters.
As shown in Figure \ref{fig:intro}, our novel Attentive Proposal Fusion (APF) module computes the similarity in features between RoI proposals and assigns higher attentive weights to similar RoIs without referring to the class labels. 
Since similar RoIs have a high likelihood of belonging to the same class, such feature representations are driven closer in the embedding space, thus forming tighter clusters. 
APF also ensures that the object detector assigns equal representation to base and novel classes, resulting in reduced catastrophic forgetting.
We also introduce a cosine margin cross-entropy loss ($L_{\text{cos-margin}}$ in Figure \ref{fig:intro}) that overcomes the impact of class confusion by increasing the angular margin between object classes. 

Existing works on FSOD demonstrate their performance on canonical benchmarks like PASCAL-VOC \cite{voc}, and MS-COCO \cite{coco} which do not represent the real-world scenarios leading to poor performance during deployment in challenging domains such as autonomous driving. On the contrary, we demonstrate the performance of our approach on the recently introduced benchmark in FSOD, few-shot India Driving Dataset \cite{majee2021fewshot} as it presents a real-world, class-imbalanced setting with large intra-class variance and inter-class bias \cite{idd}. 
\noindent The main contributions of our work can be summarized as:
\begin{itemize}
    \item We introduce a simple and lightweight metric learning based FSOD technique, Attention Guided Cosine Margin (AGCM), to overcome class confusion and catastrophic forgetting in driving scenes.
    \item We introduce a parameterless Attentive Proposal Fusion module (APF) and a Cosine Margin Cross-Entropy loss in AGCM to retain feature information from base classes while generalizing to novel classes.
    \item We demonstrate upto 10\% reduction in class confusion and 18\% improvement in catastrophic forgetting while achieving SoTA performance on the challenging India Driving Dataset (IDD) \cite{idd} and other FSOD benchmarks like PASCAL-VOC \cite{voc}.
\end{itemize}

\section{Related Work}
\label{rel}
\begin{figure*}[htbp]
    \includegraphics[width=2.2\columnwidth]{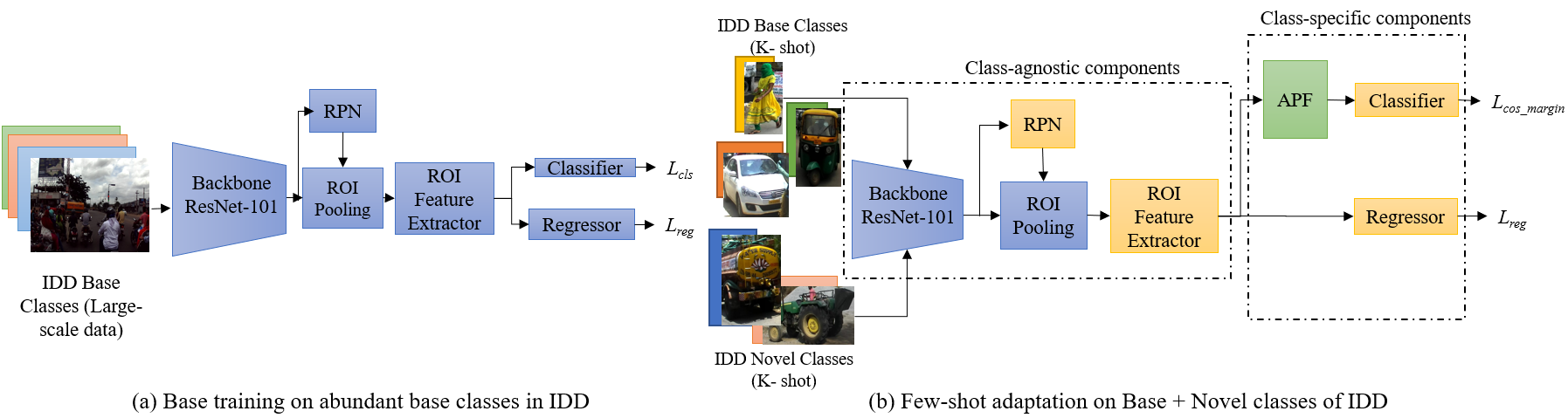}

    \caption{\textbf{The architecture of our proposed AGCM approach: } AGCM follows a metric learning strategy and is applied during the few-shot adaptation stage (b) after base training (a), on K shot samples from base and novel categories. We introduce an Attentive Proposal Fusion (APF) module and a cosine margin cross-entropy loss to overcome class confusion and catastrophic forgetting in FSOD.}
    \label{fig:blockArch}
\end{figure*}

\subsection{Few-Shot Object Detection}
\label{rel:fsod}
Classical approaches in FSOD adopt a traditional finetuning strategy \cite{lstd}, or a distance metric learner \cite{repmet} to extend the features of the already learned (base) classes to the newly added (novel) classes.
Recent approaches in FSOD adopt meta-learning techniques on standard object detection methods using episodic training \cite{reweight, addfeat, metarcnn} to learn class-specific feature sets to discriminate among classes. 
Meta-Reweight \cite{reweight} and MetaRCNN \cite{metarcnn} learns an additional feature extractor network that converts class agnostic features from the RoI head of the object detector to class-specific features.
Add-Info \cite{addfeat} introduces the feature difference between meta-train (support) and meta-test (query) images as additional features, while
\cite{fsod} learns an Attention-based Region Proposal Network (RPN) to guide a relation network \cite{relation-net} to learn discriminative features for each class. 
Very recent approaches in meta-learning like \cite{mutualsupport} encourages sharing of information between support and query images to enhance class-specific feature sets 
While CME \cite{cme} establishes an equilibrium between class margins to reduce class confusion and demonstrates better generalization to novel classes. 
A characteristic feature of meta learners is the use of attention mechanisms \cite{fsod, Zhang_2020_ACCV} to identify the most discriminative features for each class. This allows meta learners to retain the knowledge of base classes while generalizing to novel classes.

Despite their success in retaining the knowledge of base classes, meta-learning approaches are compute and memory-intensive. Surprisingly, metric learning strategies provide a better generalization to novel objects without any additional overheads. FsDet \cite{fsdet} learn generalizable feature embeddings by introducing a cosine-similarity-based classifier.  
FSCE \cite{fscontrastive} adopts a contrastive training strategy while SRR-FSD \cite{srr} uses the semantic relationships between word embeddings from category labels to show improvements on novel class performance. 
PNPDet \cite{pnpdet} decouples the base and novel class predictors and learns a cosine-similarity classifier that partially resolves catastrophic forgetting and class confusion. 
Unfortunately, metric learners suffer from extreme catastrophic forgetting as they tend to overfit on the novel classes.
GFSD \cite{Fan_2021_CVPR} proposes a Bias-Balanced RPN to prevent overfitting on metric learners and introduces a Re-detector network that decouples the base and novel class predictors. Although this technique reduces the impact of catastrophic forgetting, it fails to generalize to novel classes.
Our work introduces attention-based proposal level fusion into metric learning based FSOD technique to help retain information from base classes, preventing catastrophic forgetting. 

The authors in \cite{majee2021fewshot} demonstrate the application of FSOD in the context of autonomous driving to detect less-occurring road objects. Our work adopts this problem definition and demonstrates a reduction in catastrophic forgetting while showing significant improvements in the performance of novel classes.

\subsection{Margin based Feature learning}
\label{}
Margin-based learning has been applied to various computer vision tasks \cite{large-margin, mettes2019hyperspherical, face} to better discriminate between objects that show a significant overlap in visual features. 
Such penalties have proven to be effective in reducing class confusion for the few-shot classification \cite{Li_2020_CVPR, negativemargin} task by introducing an additional angular margin between feature clusters.
While similar approaches have been recently adopted in meta-learning based FSOD techniques \cite{cme, mutualsupport}, the margin-based penalty is yet to be explored for metric learners.
To the best of our knowledge, we are the first to introduce a simple and effective margin-based penalty in metric learning based FSOD techniques through the Cosine Margin Cross-Entropy loss described in section \ref{meth:cos}.

\section{Method}
\label{meth}
In this section, we define the problem for Few-Shot Object Detection and describe the architecture of our proposed Attention Guided Cosine Margin (AGCM) approach.

\subsection{Problem Definition}
\label{meth:prob_def}
We define a proposal based few-shot object detector $h(I,\theta)$ consisting a class-agnostic component $f(I,\theta_{f})$ and a class-specific component $C(f,\theta_{c})$ as shown in figure \ref{fig:blockArch}(b), such that $h(I,\theta) = C(f(I,\theta_{f}),\theta_{c})$. 
Here, $I$ represents the input images and $\theta$, $\theta_{f}$ and $\theta_{c}$ represents the respective model parameters for the components of the few-shot object detector. 
We define a metric learning based FSOD training strategy as in \cite{majee2021fewshot} which proceeds in two stages: \emph{base training} and \emph{few-shot adaptation}. During base training $h(I,\theta)$ learns to detect objects from base classes ($C_{base})$ using a large scale dataset $D_{base}$. In the few-shot adaptation stage, $h(I, \theta)$ is fine-tuned using images in $D_{novel}$ consisting of classes $C_{base} \cup C_{novel}$ with only $K$ instances from $N$ classes, such that $| C_{base} \cup C_{novel} | = N$.
The goal for $h(I,\theta)$ is to boost performance on novel classes in $D_{novel}$ with minimal degradation in performance of classes in $D_{base}$.

\subsection{AGCM: Attention Guided Cosine Margin}
\label{meth:fsmar}
The proposed Attention Guided Cosine Margin (AGCM) approach adopts a novel metric learning strategy to reduce the intra-class variance and inter-class bias among object classes by encouraging orthogonality among class-specific feature clusters \cite{opl}.
As shown in Figure \ref{fig:blockArch}(b) the AGCM is applied only during the few-shot adaptation stage to the output of the class-agnostic branch of the object detector $f(I,\theta_{f})$ to guide the class-specific component $C(f,\theta_{c})$ through two key components. We first apply a novel Attentive Proposal Fusion (APF) to the feature representations of individual RoI proposals in $P = f(I,\theta_{f})$. 
The feature information in each RoI proposal $P_{i}$ is propagated across all proposals $P_{j} \in P$ and is fused with those that have high visual similarity with $P_{i}$ in a label-free fashion.
Secondly, we introduce a Cosine Margin Cross-Entropy loss term to the classification head of the object detector $C(f,\theta_{c})$. This loss term maximizes the angular separation between feature clusters to reduce inter-class bias among classes. We describe the formulations of the APF and Cosine Margin Cross-Entropy loss in sections \ref{meth:frm} and \ref{meth:cos} respectively.
The combined effect of these two modules results in a significant reduction in class confusion and catastrophic forgetting, as shown by our experiments in section \ref{exp}.

\subsubsection{APF: Attentive Proposal Fusion Module}
\label{meth:frm}
The scarcity of data samples in FSOD techniques leads to the formation of non-discriminative feature sets, especially for the novel classes. 
The bias associated with few-shot data has been identified in \cite{parameterless} as \emph{sample bias} for image recognition tasks. The authors in \cite{parameterless} propose a transductive meta-learning strategy by propagating feature information between labeled few-shot samples (support set) and unlabelled test data samples (query set). We adopt a similar direction through the Attentive Proposal Fusion (APF) with modifications towards metric learning based FSOD techniques.


The class-agnostic component of the object detector $f(I,\theta_{f})$ produces feature representations from $M$ RoI proposals, denoted by $P$.
Our proposed APF module is applied to individual RoI proposals in $P$ to maximize the class-specific feature information by fusing the low-level features from RoI proposals $p_{i} \in P$ with the weighted sum of the remaining $M-1$ proposals as described in equation \ref{eq:rerepresent}. 
Here, $\Phi\left(p_{i}\right)$ represents the proposal $p_{i}$ after feature fusion and $w_{ij}$ is the attentive weight between the $i_{th}$ and the $j_{th}$ RoI proposal.
\begin{equation}
\label{eq:rerepresent}
\Phi\left(p_{i}\right) = \alpha \cdot p_{i}+ (1-\alpha)\sum_{j \in \mathbb{P}, j \neq i} w_{ij} \cdot p_{j}
\end{equation}
As described in equation \ref{eq:attn} the attentive weights ($w_{ij}$) represents the likelihood of the features in the $i_{th}$ RoI proposal to be similar to the features in the $j_{th}$ proposal. It involves a non-linear similarity (cosine similarity \cite{nonforget} in our case) between $p_{i}$ and $p_{j}$ denoted by $cos(p_{i},p_{j})$. The choice of this metric is described in detail in section \ref{sec:hyper_cos}.
\begin{equation}
\label{eq:attn}
w_{ij}=\frac{e^{cos(p_{i}, p{j})}}{\sum_{k \neq i, k \in \mathbb{P}} e ^{cos(p_{i}, p{k})}}
\end{equation}
The formulation of $\Phi\left(p_{i}\right)$ introduces a hyper-parameter $\alpha$ which controls the proportion of low-level features that are fused into $p_{i}$ from remaining RoI proposals. The value of $\alpha$ is always kept in the range $[0.5,1.0]$ to encourage the retention of a significant portion of the features of the original RoI. More details on the choice of $\alpha$ is provided in section \ref{sec:hyper_apf}.

The information exchange among RoI proposals encourages the grouping of similar feature representations without the ground truth label information. 
This facilitates the reduction in intra-class bias and chances of model overfitting as all classes in the training dataset are equally represented in the embedding space. Consequently, we observe a diminishing effect on catastrophic forgetting of base classes as shown in section \ref{abl:baseforget}.

\subsubsection{Cosine Margin Cross Entropy Loss}
\label{meth:cos}
Although the application of label-free feature fusion (APF module) helps in forming tighter feature clusters, it may result in the clustering of features from heterogeneous classes that show high visual similarities. It also fails to ensure sufficient margin among co-occurring object classes like \emph{motorcycle} and \emph{rider}, leading to elevated class confusion. 

Based on the recent success of margin based penalties in auxiliary vision tasks (section \ref{}), we introduce 
a negative angular margin based loss function in AGCM with suitable modifications for metric learning based FSOD techniques. 
In contrast to a positive margin, a negative margin helps establish an equilibrium between the distinguishability of classes and the performance of novel classes \cite{negativemargin}. 

We apply the cosine margin-based objective in the few-shot adaptation stage, to the output logits of the classification head in the FSOD model, $Z = C(\phi,\theta_{c})$, where $\theta_{c}$ represents the parameters of the classifier head and $\phi$ is obtained by applying APF module on the RoI features from the class-agnostic branch $f(I,\theta_{f})$. 
The objective function described in equation \ref{eq:lmargin} through \ref{eq:margin_z} maximizes the log-likelihood of the angular distance between the logit $z_{i} \in Z$ corresponding to the ground truth label $y_{i}$ and the normalized weight vector of the corresponding class $W_{y_{i}}$. 
\begin{equation}
\label{eq:lmargin}
L_{\text{cos-margin}} = -\frac{1}{M} \sum_{i=1}^{M} l(z_i) 
\end{equation}

\begin{equation}
\label{eq:margin_z}
l_{z_i} = \log \frac{e^{\beta \left(\mathrm{cos}\left(z_{i}, W_{y_{i}}\right)- \mathbbm{1}_{y_i \neq back} m\right)}}{e^{\beta \left(\mathrm{cos}\left(z_{i}, W_{y_{i}}\right)- \mathbbm{1}_{y_i \neq back} m\right)}+\sum_{j=1, j \neq y_{i}}^{N} e^{\beta \mathrm{cos}\left(z_{i}, W_{j}\right)}}
\end{equation}
An angular margin $m$ is applied to this objective to increase the separation between feature clusters, and a scaling factor $\beta$ is introduced which is set to a constant value of 20 \cite{negativemargin}. 
Also, we do not apply the angular margin to logits of the \emph{background} class to prevent loss of information during model training as the \emph{background} class proposals might contain features belonging to one or more object classes in $C_{base} \cup C_{novel}$.
The choice of the value of margin $m$ is described in section \ref{sec:hyper_cos}.

\begin{table*}[h]
      \caption{\textbf{Results on Few-Shot India Driving Dataset: }Few-shot object detection performance ($mAP_{50}$) on IDD-OS and IDD-10 splits from India Driving Dataset using 5 and 10 shot samples. }
      \centering
      \scalebox{0.8}{
      \begin{tabular}{|l|cc|cc|cc|cc|cc|cc|}
            \hline
            \textbf{Data-split}          & 
            \multicolumn{4}{c}{\textbf{IDD-OS}} &
            \multicolumn{4}{c|}{\textbf{IDD-10 (Split 1)}}     &   \multicolumn{4}{c|}{\textbf{IDD-10 (Split 2)}}  \\ 
            \hline
            \textbf{Shots (K)}          &  \multicolumn{2}{c|}{\emph{K=5}}     &   \multicolumn{2}{c|}{\emph{K=10}}  &  \multicolumn{2}{c|}{\emph{K=5}}     &   \multicolumn{2}{c|}{\emph{K=10}} &  \multicolumn{2}{c|}{\emph{K=5}}     &   \multicolumn{2}{c|}{\emph{K=10}}\\ 
            \hline
            \backslashbox{\textbf{Method}}{\textbf{Metric}}         & $mAP_{base}$ & $mAP_{novel}$ & $mAP_{base}$ & $mAP_{novel}$ & $mAP_{base}$ & $mAP_{novel}$ & $mAP_{base}$ & $mAP_{novel}$ & $mAP_{base}$ & $mAP_{novel}$ & $mAP_{base}$ & $mAP_{novel}$\\
            \hline
            Meta-RCNN  \cite{metarcnn}     &    24.1  & 4.3  & 24.0  & 6.4    & 23.2  & 5.7  & 24.6  &  7.8 & 18.1  & 7.4  & 18.2  & 6.7      \\
            Add-Info \cite{addfeat} &  36.4  & 18.2  & 37.1 & 28.8  & 33.5  & 5.2  & 33.7      & 10.0  & 31.3  & 7.7  & 32.1 & 9.5    \\
            FsDet w/ cos \cite{fsdet}  &  38.2  & 23.6  & 47.8 & 39.8   & 33.5  & 13.1  & 31.2  & 22.1  & 34.2  & 14.8  & 39.7  & 22.8   \\
            FSCE \cite{fscontrastive}         &   38.1	& 39.1	& 45.5 &	51.6 &	23.6	& 9.2	& 31.3 &	16.4 &	30.6 &	9.1 &	37.7 &	14.7   \\
            \hline
            \textbf{AGCM (ours)} &  \textbf{42.1} &	\textbf{45.5} &	\textbf{51.5} &	\textbf{58.0} &	\textbf{37.2} &	\textbf{16.0} &	\textbf{45.0}	 & \textbf{22.1} &	\textbf{36.2} &	\textbf{15.2} &	\textbf{42.3} &	\textbf{24.8}   \\
            \hline
      \end{tabular}}
      \label{tab:idd_os}
\end{table*}
\newcommand{\centered}[1]{\begin{tabular}{l} #1 \end{tabular}}

\begin{figure*}
        \centering
        \begin{tabular}{lcccc}
            \rotatebox{90}{FSCE \cite{fscontrastive}} &
                \includegraphics[width=0.23\textwidth]{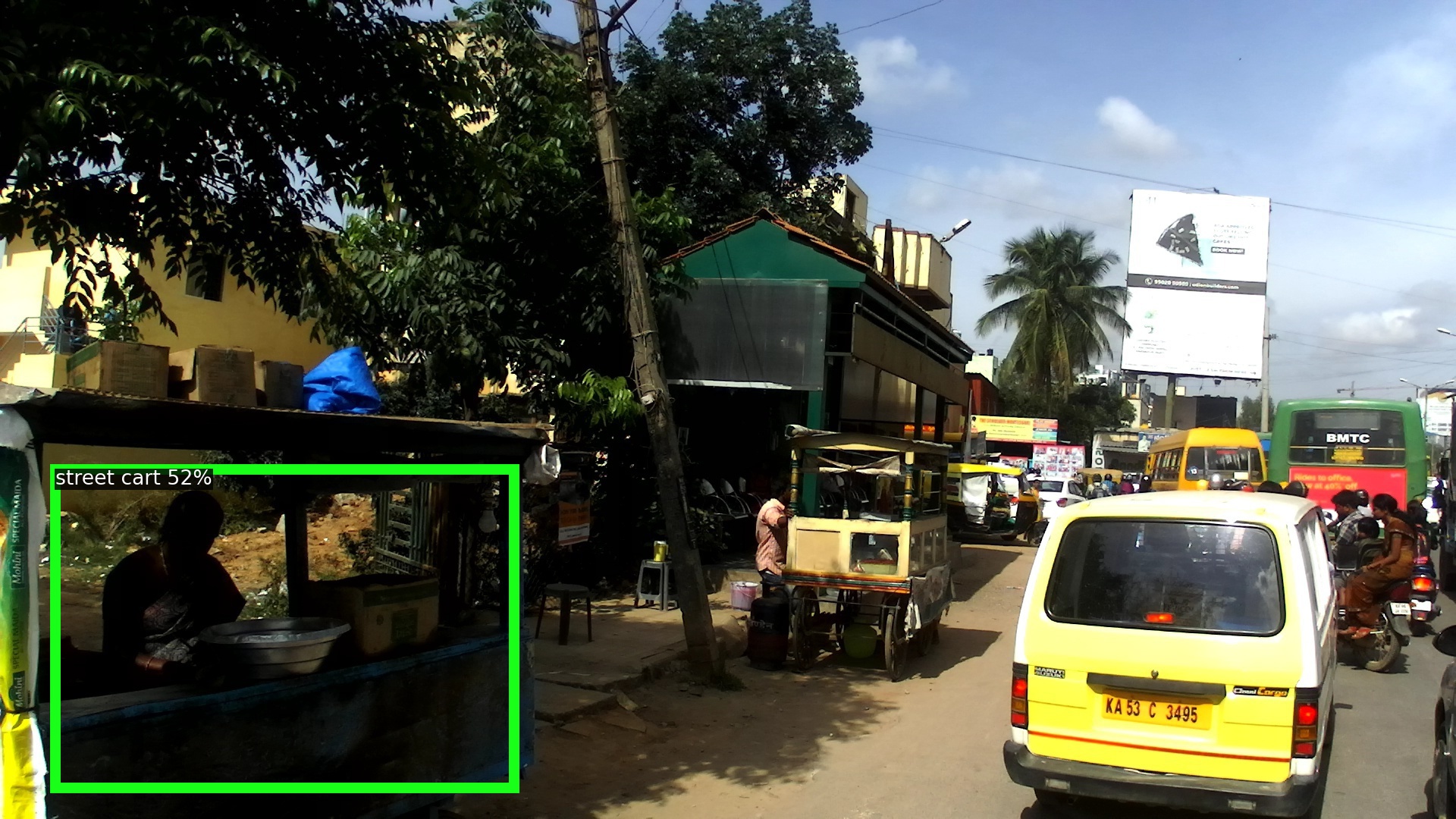} & \includegraphics[width=0.23\textwidth]{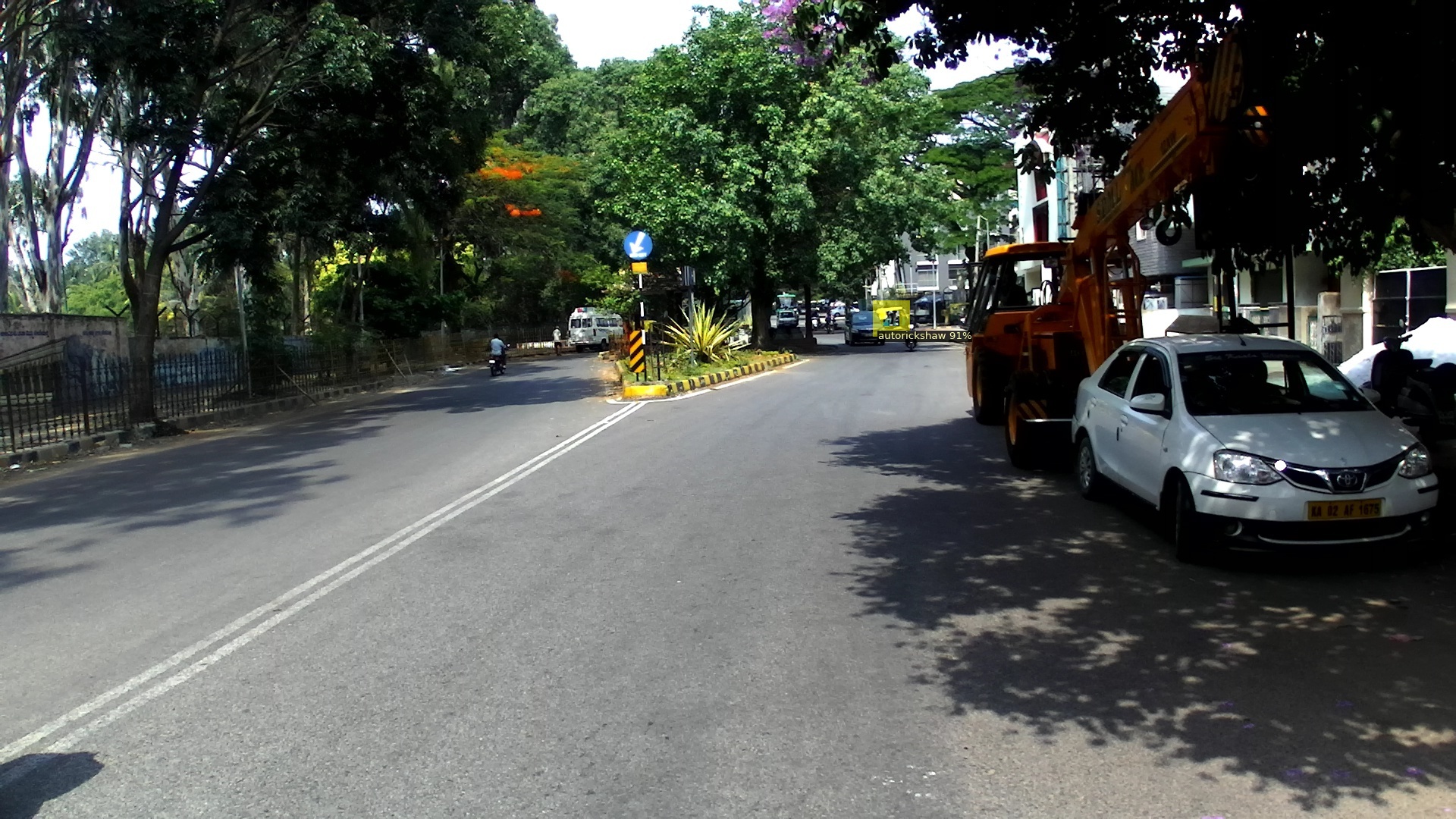} & \includegraphics[width=0.23\textwidth]{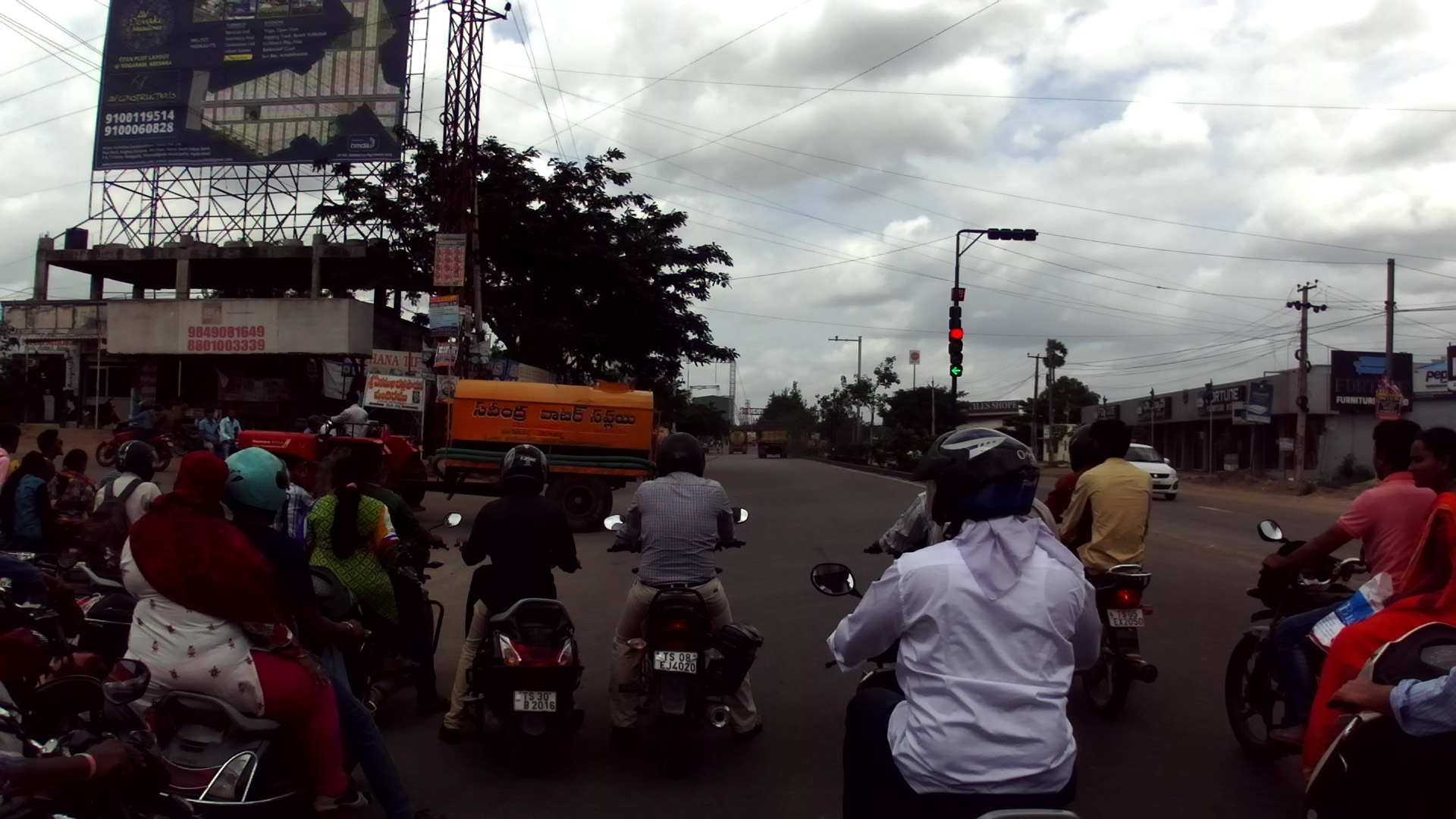} & \includegraphics[width=0.23\textwidth]{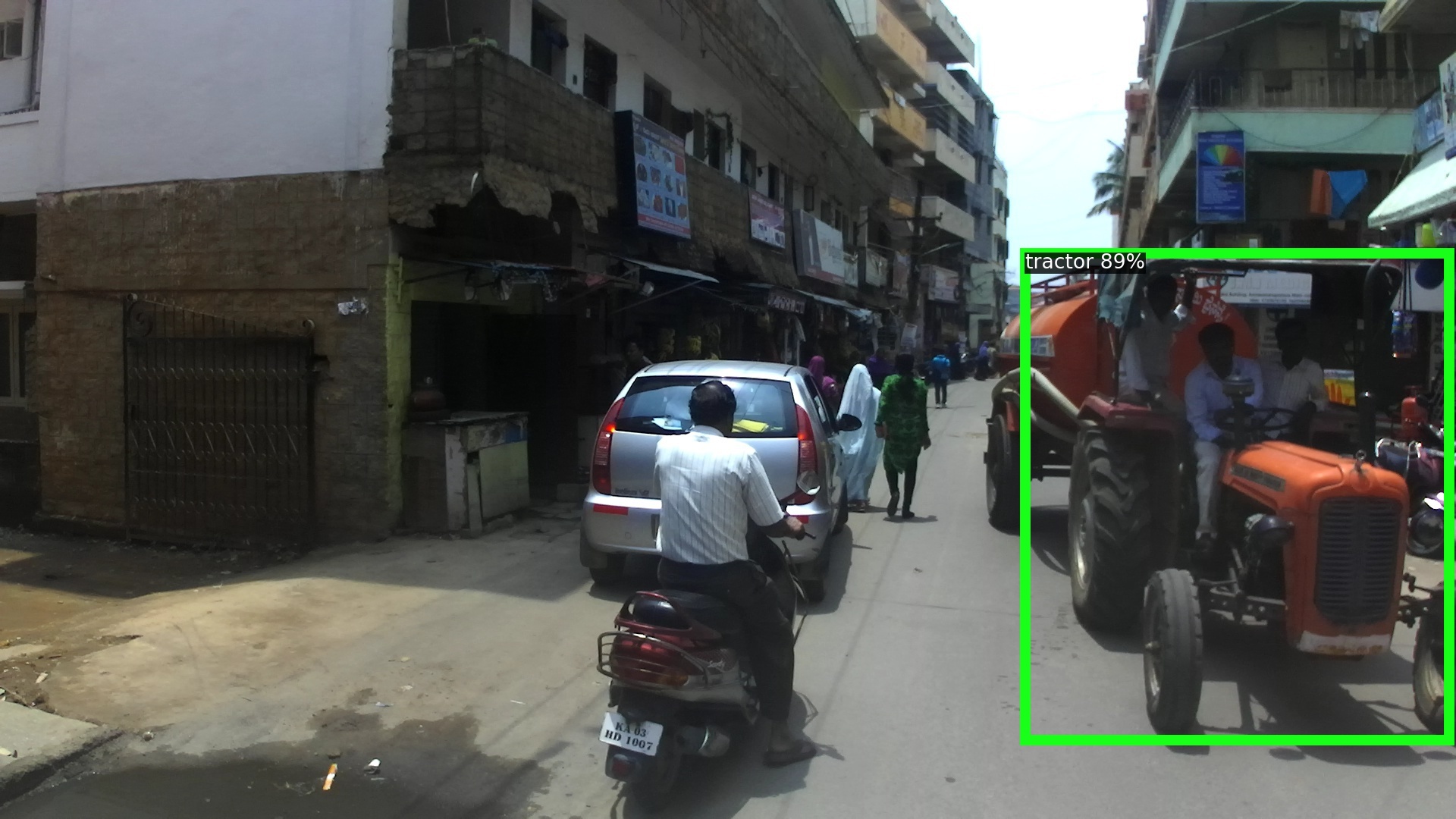} \\
            \rotatebox{90}{AGCM (ours)} & 
                \includegraphics[width=0.23\textwidth]{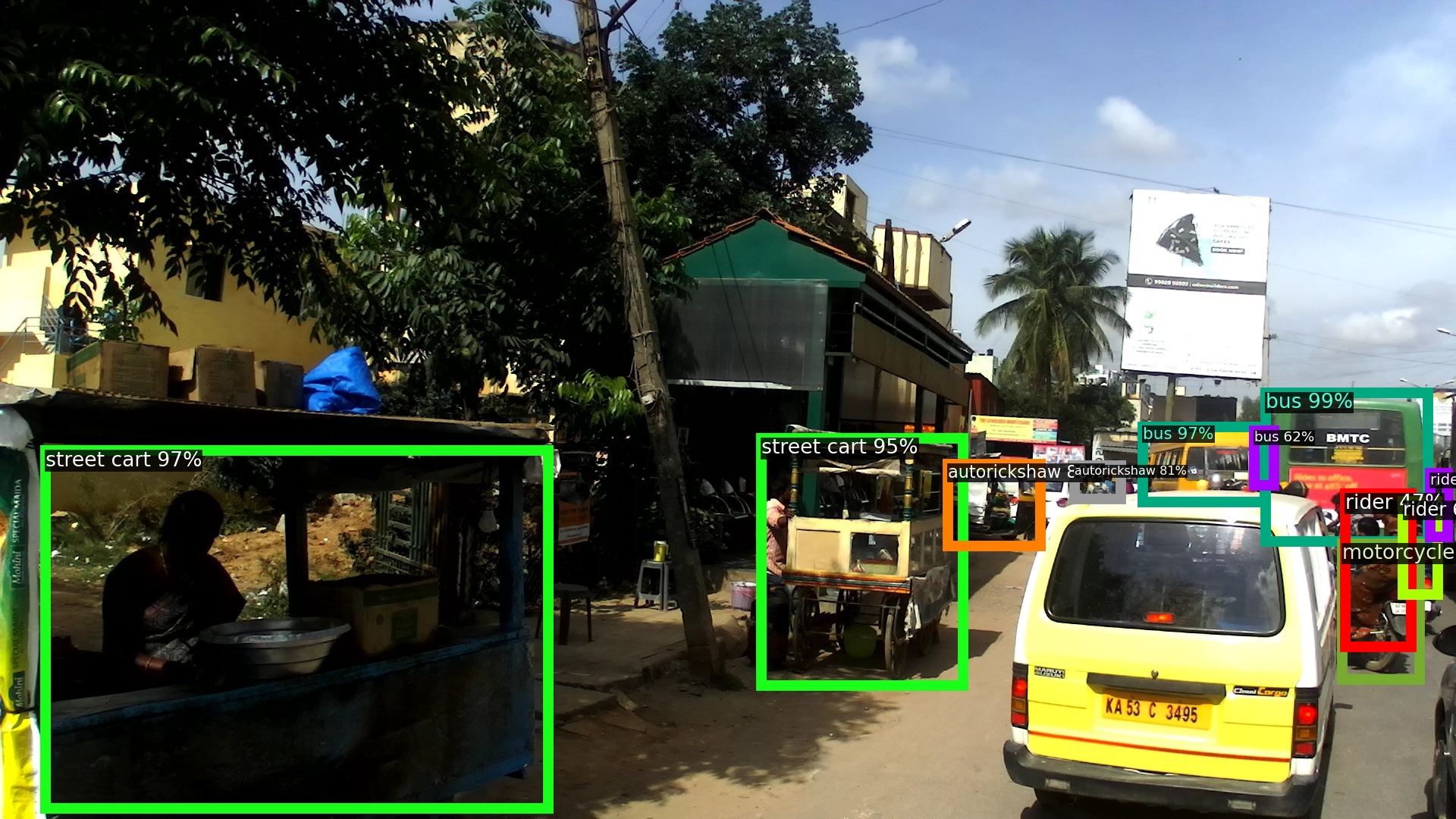} & \includegraphics[width=0.23\textwidth]{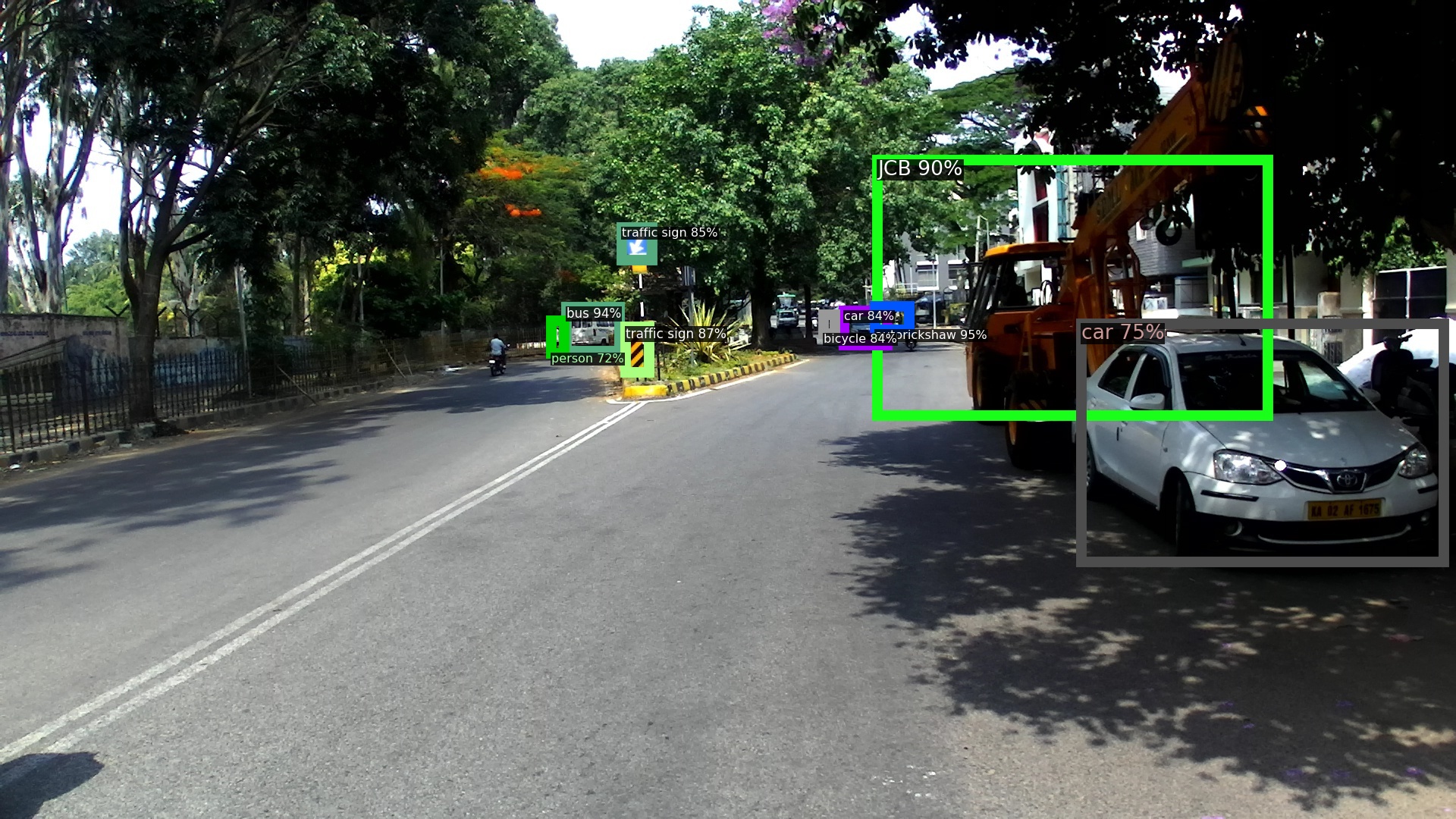} & \includegraphics[width=0.23\textwidth]{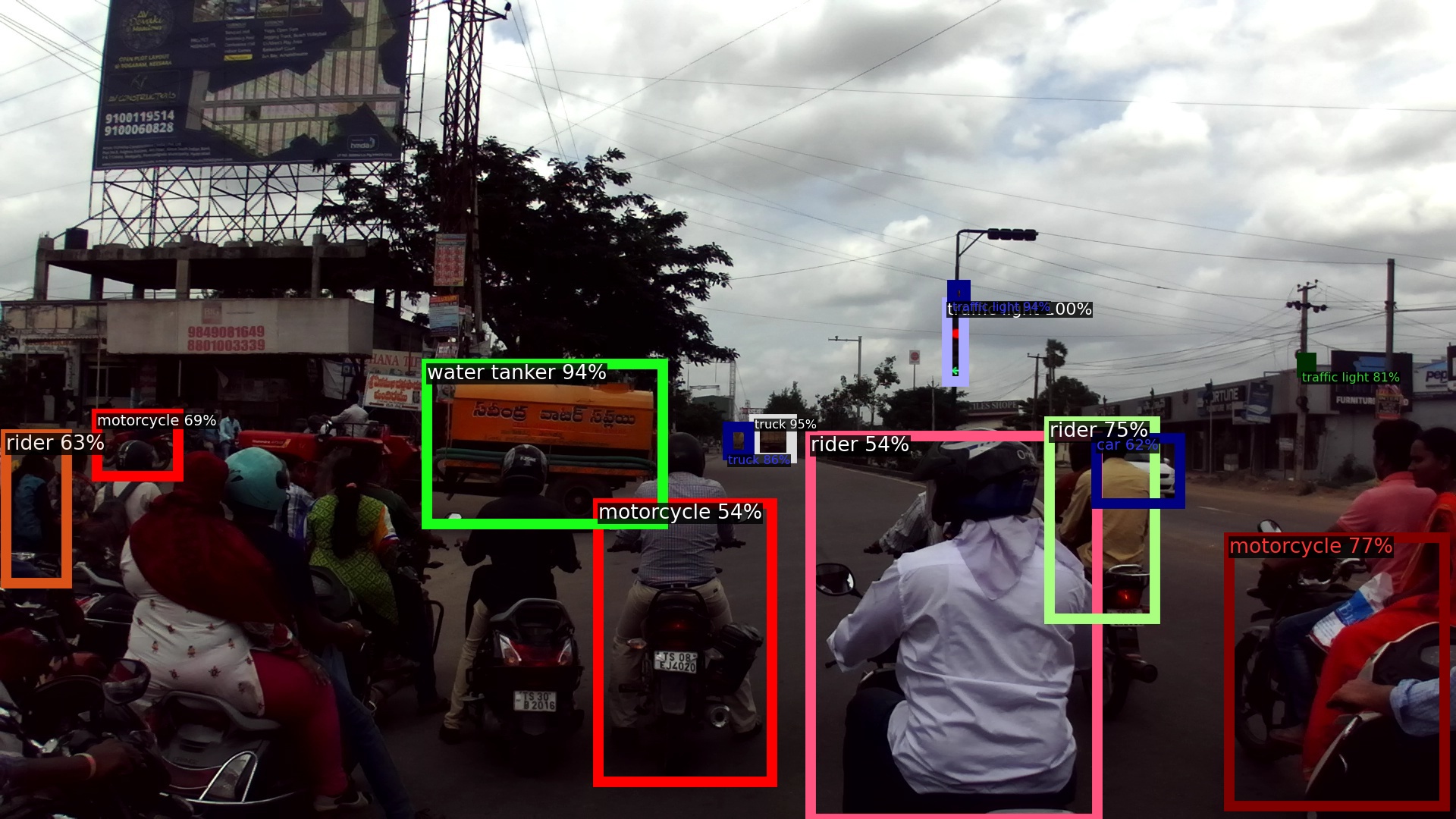} & \includegraphics[width=0.23\textwidth]{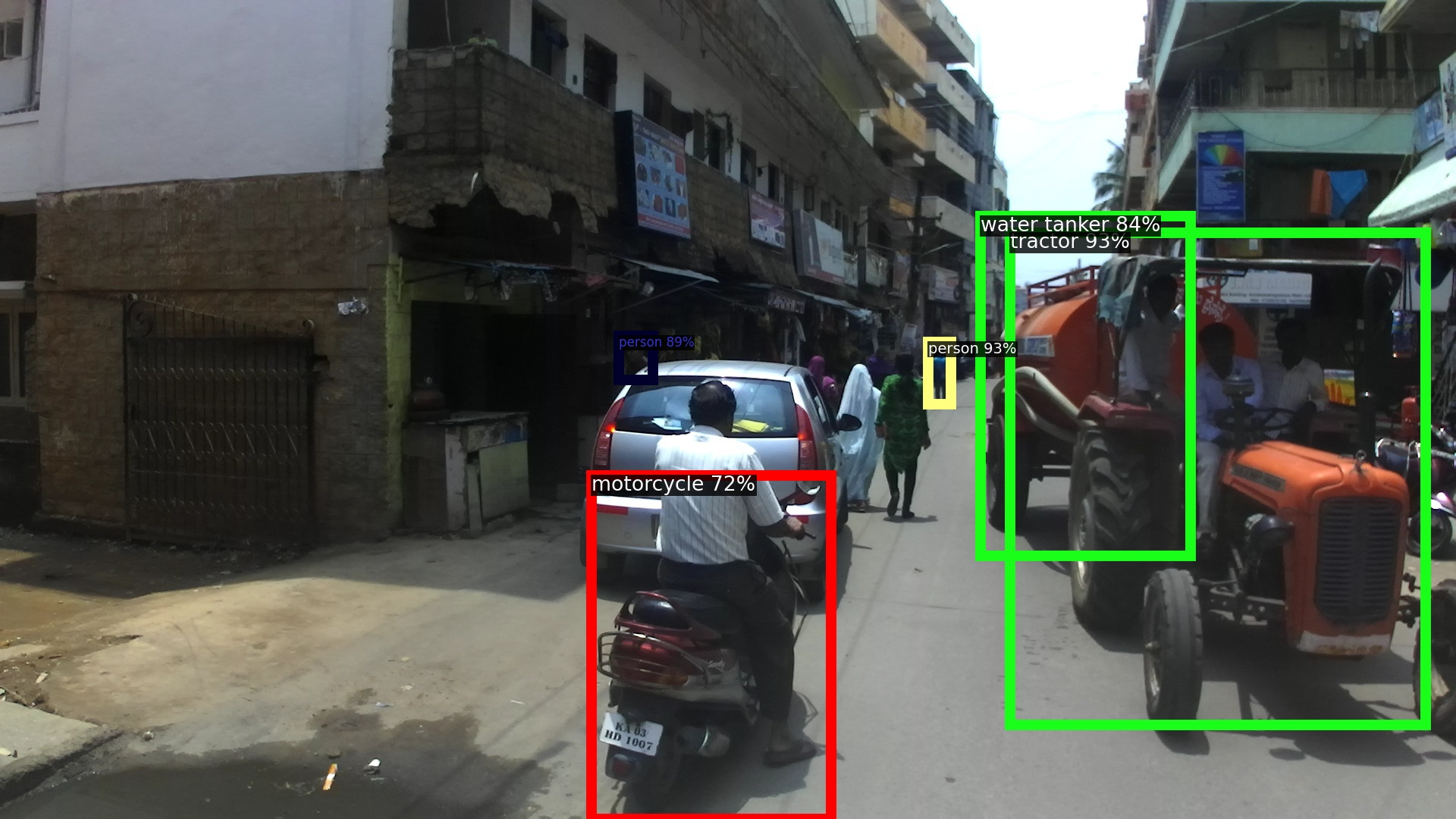} \\
            & (a) Street Cart & (b) Excavator & (c) Water Tanker & (d) Tractor \\
        \end{tabular}
        \caption{\textbf{Qualitative results from the few-shot India Driving Dataset: } We contrast the performance of AGCM against SoTA FSOD approach, FSCE for novel classes in the IDD-OS split for the 10-shot setting. FSCE suffers from extreme catastrophic forgetting and is unable to adapt to large intra-class and inter-class variations in IDD. Such issues are shown to have been overcome by the proposed AGCM approach. }
        \label{fig:qual}
\end{figure*}

\subsubsection{Training Procedure}
\label{meth:training_procedure}
As defined in section \ref{meth:prob_def} the model $h(I,\theta)$, involves a Faster-RCNN \cite{faster-rcnn} based object detector and trained in two distinct stages. We adopt the training strategy of FsDet \cite{fsdet} during the base training stage, and train $h(I,\theta)$ till convergence. We use the standard loss function used in \cite{faster-rcnn} comprising of a binary cross-entropy loss at the Region Proposal Network (RPN) to separate foreground and background proposals $L_{rpn}$, a cross-entropy loss for bounding box classifier $L_{cls}$ and a smoothed L1 loss to localize the bounding box deltas $L_{reg}$. 

In the few-shot adaptation stage, we adopt the stronger baseline presented by FSCE \cite{fscontrastive} in which the network backbone remains frozen, the number of proposals generated by the RPN is doubled, and the number of RoI features used for loss computation is halved. 
This is done to facilitate the incorporation of the low-confidence predictions from the novel classes during the initial training iterations. 
We add our APF module to the classifier head of $h(I,\theta)$ and replace the cross-entropy loss $L_{cls}$ with our proposed cosine margin cross-entropy loss $L_\text{cos-margin}$ as shown in \ref{eq:loss}.

\begin{equation}
\label{eq:loss}
    L = L_{rpn} + L_\text{cos-margin} + L_{reg}
\end{equation}

\section{Experiments}
\label{exp}
In this section, we describe our experimental setup and compare the results of our proposed method with existing FSOD techniques on multiple benchmark datasets. We adopt the standard evaluation criterion in FSOD \cite{fsdet, reweight} and report the Mean Average Precision ($mAP$) at 50\% Intersection Over Union (IoU) for all our experiments.

\subsection{Datasets}
\label{exp:dataset}
We evaluate our proposed AGCM approach on two few-shot object detection datasets - India Driving Dataset (IDD) \cite{idd} and PASCAL-VOC \cite{voc} datasets. 

\textbf{Indian Driving Dataset (IDD)} comprises of 15 object classes in the IDD-Detection dataset consisting of driving scenes on Indian roads. We adopt the data splits proposed in \cite{majee2021fewshot} to evaluate our AGCM approach. The dataset consists of two few-shot data splits -
\begin{itemize}
    \item \textbf{IDD-OS} consists of 14 classes representing an open-world deployment setting with 10 base classes and 4 novel classes. The novel classes (\emph{Tractor}, \emph{Street Cart}, \emph{Water tanker} and \emph{Excavator (JCB)}) have been obtained by expanding on the \emph{vehicle fallback} category in IDD. 
    \item \textbf{IDD-10} consists of 10 classes forming 2 few-shot data splits. Each split consists of 7 base classes and 3 randomly chosen novel classes. The authors of \cite{majee2021fewshot} create two representative splits, referred to as split 1 (\emph{bicycle}, \emph{bus} and \emph{truck} / others) and split 2 (\emph{auto-rickshaw}, \emph{motorcycle}, \emph{truck}/ others) based on the choice of novel classes. We adopt these splits in our work.
\end{itemize}
We evaluate our approach on the complete validation set of IDD for 5 and 10 shot settings.

\textbf{PASCAL-VOC} \cite{voc} dataset consists of 20 classes, out of which 15 are considered as base and 5 as novel classes. The novel classes are chosen at random giving rise to three data splits namely, split-1 (\emph{bird}, \emph{bus}, \emph{cow}, \emph{motorbike}, \emph{sofa}), split-2 (\emph{aeroplane}, \emph{bottle}, \emph{cow}, \emph{horse}, \emph{sofa}) and split-3 (\emph{boat}, \emph{cat}, \emph{motorbike}, \emph{sheep}, \emph{sofa}). Following previous works \cite{reweight}, we use the combined VOC 07+12 datasets for training and evaluate our models on the complete validation set of VOC 2007 for 1, 5, and 10 shot settings. 

\subsection{Experimental Setup}
\label{exp:setup}
The architecture of the proposed AGCM is based on the Faster-RCNN \cite{faster-rcnn} model with a ResNet-101 \cite{resnet} and Feature Pyramidal Network \cite{fpn} based backbone. 
For IDD, the input batch size to the network is set to 2 and 6 in the base training and few-shot adaptation stages respectively. However, for PASCAL-VOC, a batch size of 16 is used for both stages.
The input resolution is set to 1920 x 1080 pixels for data splits in IDD, while it is set to 800 x 600 pixels for PASCAL-VOC. 
Following the training procedure described in section \ref{meth:training_procedure}, we train our model till convergence with a learning rate of 0.001 for both base and few-shot adaptation stages.
For IDD, base-training is done for 50k iterations with a pretrained imagenet \cite{imagenet} backbone, while the training procedure of FSCE \cite{fscontrastive} is followed for PASCAL-VOC. 
Standard data augmentation like horizontal flip and random crop are applied for both datasets. 
During the few-shot adaptation stage, we adopt the stronger baseline of FSCE and set the number of RoI proposals to 2000 and the number of RPN proposals to 256. 
The hyper-parameters used in the formulation of AGCM, namely $\alpha$, margin ($m$), and distance, are chosen through ablation experiments described in section \ref{abl}.
Results from existing methods are a reproduction of the algorithm from publicly available codebases along with hyper-parameter tuning on IDD datasplits. 
Unlike other FSOD benchmarks, all our experiments are performed on a single GPU with 12GB memory. 

\begin{table*}[t]
      \caption{\textbf{Quantitative analysis on PASCAL-VOC dataset:} Few-shot object detection performance ($mAP_{novel}$) on novel class splits of PASCAL-VOC dataset. We tabulate results for K={1, 5, 10} shots from various SoTA techniques in FSOD. * indicates that the results are averaged over 10 random seeds. $\dagger$ indicates a different evaluation strategy (N-way, K-shot meta testing). }
      \centering
      \small
      \begin{tabular}{|l|c|c|ccc|ccc|ccc|}
            \hline
            \textbf{Method}          & \multicolumn{1}{p{2cm}|}{\centering Meta/ Metric \\ Learner} & Backbone &  
                                        \multicolumn{3}{c|}{\textbf{Novel Split 1}} &   
                                        \multicolumn{3}{c|}{\textbf{Novel Split 2}} &
                                        \multicolumn{3}{c|}{\textbf{Novel Split 3}} \\ \cline{4-12}
                                     &&&  K=1 & 5 & 10 &
                                          1   & 5 & 10 &
                                        1   & 5 & 10 \\ 
            \hline
            $\small{\dagger}$ Meta-RCNN \cite{metarcnn}       & Meta      & FRCN-R101 & 19.9  & 45.7  & 51.5  & 10.4  & 34.8  & 45.4  & 14.3  & 41.2  & 48.1  \\
            $\small{\dagger}$Meta-Reweight \cite{reweight}   & Meta      & YOLO V2   & 14.8  & 33.9  & 47.2  & 15.7  & 30.1  & 40.5  & 21.3  & 42.8  & 45.9  \\
            $\small{\dagger}$MetaDet \cite{metadet}          & Meta      & FRCN-R101 & 18.9  & 36.8  & 49.6  & 21.8  & 31.7  & 43.0  & 20.6  & 43.9  & 44.1  \\
            $\small{\dagger}$Add-Info \cite{addfeat}         & Meta      & FRCN-R101 & 24.2  & 49.1  & 57.4  & 21.6  & 37.0  & 45.7  & 21.2  & 43.8  & 49.6  \\
            $\small{\dagger}$CME \cite{cme}            & Meta & YOLO V2 & 17.8   & 44.8  &  47.5 &  12.7 & 33.7   & 40.0   & 15.7  &  44.9 & 48.8  \\
            \hline
            PNPDet \cite{pnpdet}            & Metric    & DLA-34    & 18.2  & - & 41.0  & 16.6  & -  & 36.4  & 18.9  & -  & 36.2  \\
            FsDet w/ FC \cite{fsdet}        & Metric    & FRCN-R101 & 36.8  & 55.7  & 57.0  & 18.2  & 35.5  & 39.0  & 27.7  & 48.7  & 50.2  \\
            FsDet w/ cos \cite{fsdet}       & Metric    & FRCN-R101 & 39.8  & 55.7  & 56.0  & 23.5  & 35.1  & 39.1  & 30.8  & 49.5  & 49.8  \\
            FSCE \cite{fscontrastive}            & Metric & FRCN-R101 & \textbf{41.0}
  &  57.9 & 57.8 & 27.3	& 44.4 & 49.8 & 40.1	& 53.2	& 57.7
   \\
            \textbf{AGCM (ours)}    & Metric & FRCN-R101 & 40.3 &	\textbf{58.5} &	\textbf{59.9} &	\textbf{27.5} &	\textbf{49.3} &	\textbf{50.6} &	\textbf{42.1} &	\textbf{54.2} &	\textbf{58.2}
  \\
            \hline 
            $\small{*}$FsDet w/ cos \cite{fsdet}       & Metric    & FRCN-R101 & 25.3  & 47.9  & 52.8  & \textbf{18.3}  & 34.1  & 39.5  & 17.9  & 40.8  & 45.6  \\
            $\small{*}$FSCE \cite{fscontrastive}            & Metric & FRCN-R101 & 28.2
  &  46.2 & 54.1 & 16.5	& 35.9 & 45.3 & 22.2	& 45.4	& 49.4 \\
        \textbf{ $\small{*}$ AGCM (ours)}    & Metric & FRCN-R101 & \textbf{28.3} &	\textbf{49.0} &	\textbf{54.8} &	17.2 &	\textbf{38.5} &	\textbf{47.0} &	\textbf{22.9} &	\textbf{46.5} &	\textbf{51.5} \\ \hline
      \end{tabular}
      \label{tab:voc_main}
\end{table*}

\subsection{Results on India Driving Dataset}
\label{exp:idd}
We follow the benchmark experiments in \cite{majee2021fewshot} and compare the performance of our AGCM approach against State-of-The-Art (SoTA) meta \cite{addfeat, metarcnn}, and metric learners \cite{fsdet} on IDD-OS and IDD-10 splits. 
Additionally, we extend this benchmark be reimplementing the results of the current SoTA approach in FSOD, FSCE \cite{fscontrastive} on IDD datasplits. 
Table \ref{tab:idd_os} records both the base and novel class performance of various approaches in contrast to our AGCM appraoch. 
For IDD-10 splits, our AGCM outperforms existing SoTA methods by an average of 1.5 $mAP$ points in split-1 and 1.2 $mAP$ points in split-2 on novel classes. 
For, IDD-OS split, AGCM outperforms the SoTA metric learner, FSCE, by 6.4 and 6 $mAP$ points for the 5 and 10 shot settings, respectively. 
Alongside the significant improvements in novel class performance, our AGCM approach achieves the highest retention in base class performance, which effectively overcomes catastrophic forgetting. This is further described in section \ref{abl:baseforget}.  
Although FSCE has proven to be effective against class confusion and catastrophic forgetting for canonical datasets, there exists a large performance gap between FSCE and AGCM on the IDD datasplits. This can be attributed to the contrastive training strategy in FSCE resulting in elimination of discriminative features for confusing road objects.

\begin{table}[t]
      \caption{\small Ablation on various components of the proposed AGCM approach. }
      \centering
      \scalebox{0.8}{
        \begin{tabular}{|l|ccc|cc|}
            \hline
            \multirow{2}{*}{Method} & Stronger & APF  & Cosine Margin & \multicolumn{2}{c|}{$mAP_{novel}$} \\ 
            \cline{5-6}             
                                    & Baseline \cite{fscontrastive} & (Sec. \ref{meth:frm}) & CE loss & 5-shot & 10-shot \\
            \hline
            FsDet w/ cos            &  -            & -          & -         &   23.6  & 39.8     \\
            FSCE                    & \checkmark    &            &              &   38.7  & 51.3     \\
            \hline
            \multirow{2}{*}{AGCM (ours)} & \checkmark    & \checkmark &              &   43.3  & 54.9     \\
                                         &  \checkmark   & \checkmark & \checkmark   &   \textbf{45.5}  & \textbf{58.0}  \\
            \hline
      \end{tabular}
      }
      \label{tab:agcm_comp}
\end{table}
\begin{table}[t]
\caption{\small Ablation for the effect of key hyper-parameters ($\alpha$, distance and $m$) on novel class performance in IDD-OS. The chosen values for the AGCM approach is \underline{underlined} and associated performance values are indicated in \textbf{bold}.}
\label{tab:hyper}
\centering
\small
\scalebox{0.9}{
\begin{tabular}{|c|c|c|c|}
\hline
Parameter & Value & $mAP_{base}$ & $mAP_{novel}$ \\ \hline
\multirow{5}{2cm}{\centering $\alpha$ \\(Distance = Euclidean)}
                              & 0.5                       & 48.9                         & 44.8                          \\
                              & 0.7                       & 52.2                         & 52.9                          \\
                              & \underline{0.8}                       & \textbf{52.7}                & \textbf{53.9}                          \\
                              & 0.9                       & 52.2                         & 54.4                          \\
                              & 1.0                       & 50.5                         & 52.7                          \\ \hline
\multirow{3}{2cm}{\centering Distance \\($\alpha = 0.8$)}     
                              & Euclidean                 & 52.7                         & 53.9                          \\
                              & \underline{Cosine}                    & \textbf{52.7}                & \textbf{54.9}                          \\
                              & Pearson                   & 52.1                         & 54.8                          \\ \hline
\multirow{6}{2cm}{\centering $m$ \\(Distance = Cosine, $\alpha$ = 0.8)}
                              & 0.0                       & 52.7                         & 54.9                          \\
                              & 0.1                       & 52.0                         & 56.1                          \\
                              & \underline{0.2}                       & \textbf{51.5}                & \textbf{57.9}                          \\
                              & 0.4                       & 50.9                         & 53.8                          \\
                              & 0.8                       & 49.1                         & 40.1                          \\
                              & 1.0                       & 48.5                         & 43.3  \\ \hline                        
\end{tabular}}
\end{table}

Figure \ref{fig:qual} demonstrates a qualitative analysis of our approach against the SoTA metric learner FSCE on the IDD-OS split in the 10-shot setting. As observed from the figure, the FSCE approach suffers from significant catastrophic forgetting (as shown in figure \ref{fig:qual}(a)) and is unable to detect incomplete (Water tanker in figure \ref{fig:qual}(d)) or obscure objects (Excavator in figure \ref{fig:qual}(b)). Unlike FSCE, the AGCM approach can retain most base class predictions and is invariant to large intra-class variances in IDD.

\subsection{Results on PASCAL-VOC dataset}
\label{exp:voc}
Table \ref{tab:voc_main} records the results obtained from our AGCM approach on novel splits of the PASCAL-VOC dataset and contrasts it against SoTA FSOD techniques. 
Our method outperforms SoTA approaches on almost all few-shot settings with a maximum improvement of 4.9 $mAP$ points in split-2 for the 5-shot setting. 
However, we do not achieve high gains for very low shot settings (1-shot) as the model suffers significant inter-class bias and intra-class variance. 

\section{Ablation}
\label{abl}
In this section, we conduct ablation experiments on the challenging IDD-OS split to qualify the contributions of various components and hyper-parameters in our proposed AGCM approach. 

\begin{figure*}[tp]
        \centering
        \small
        \begin{tabular}{ccc}
                \includegraphics[width=0.30\textwidth]{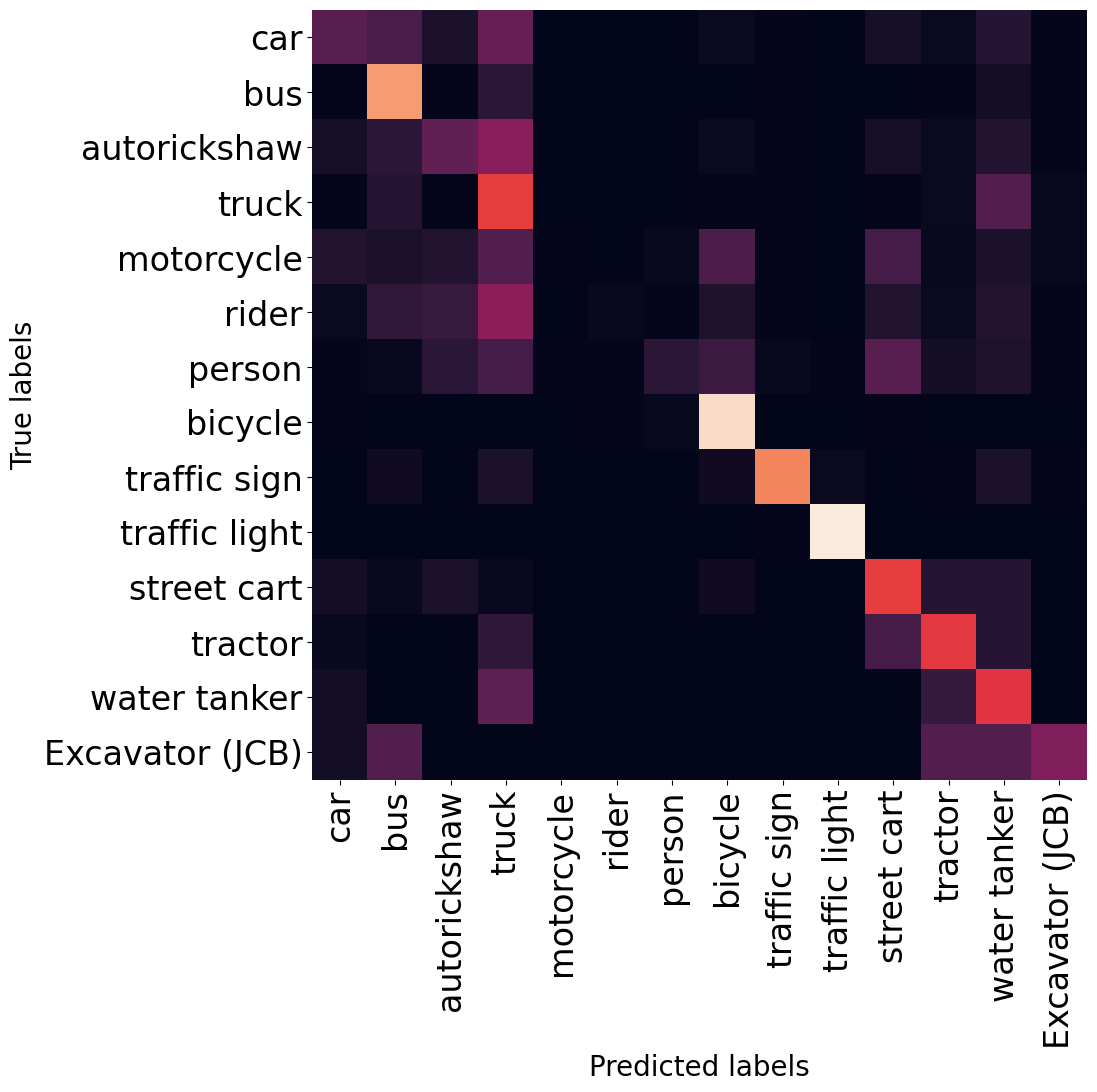} &
                \includegraphics[width=0.30\textwidth]{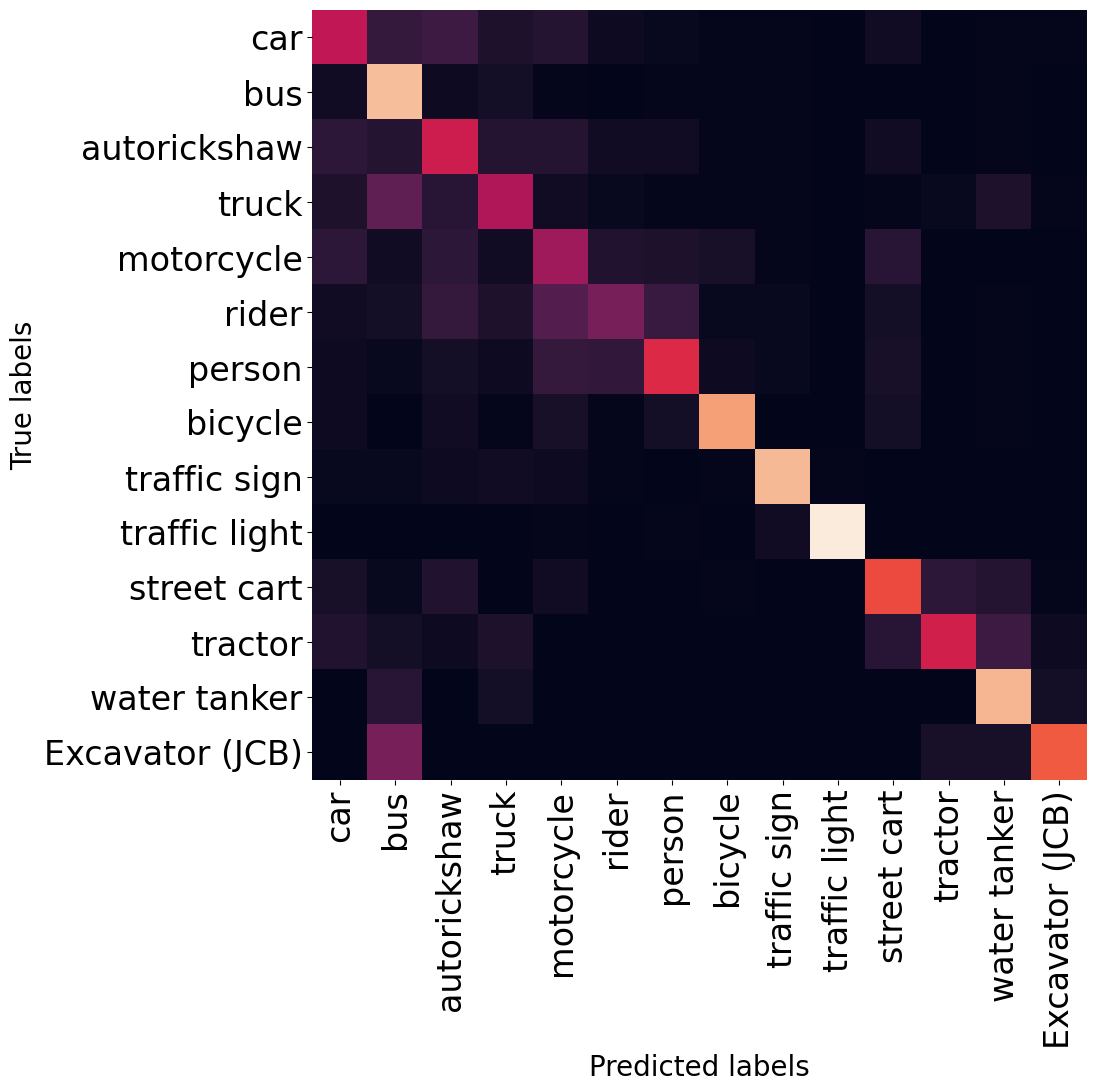} & 
                 \includegraphics[width=0.30\textwidth]{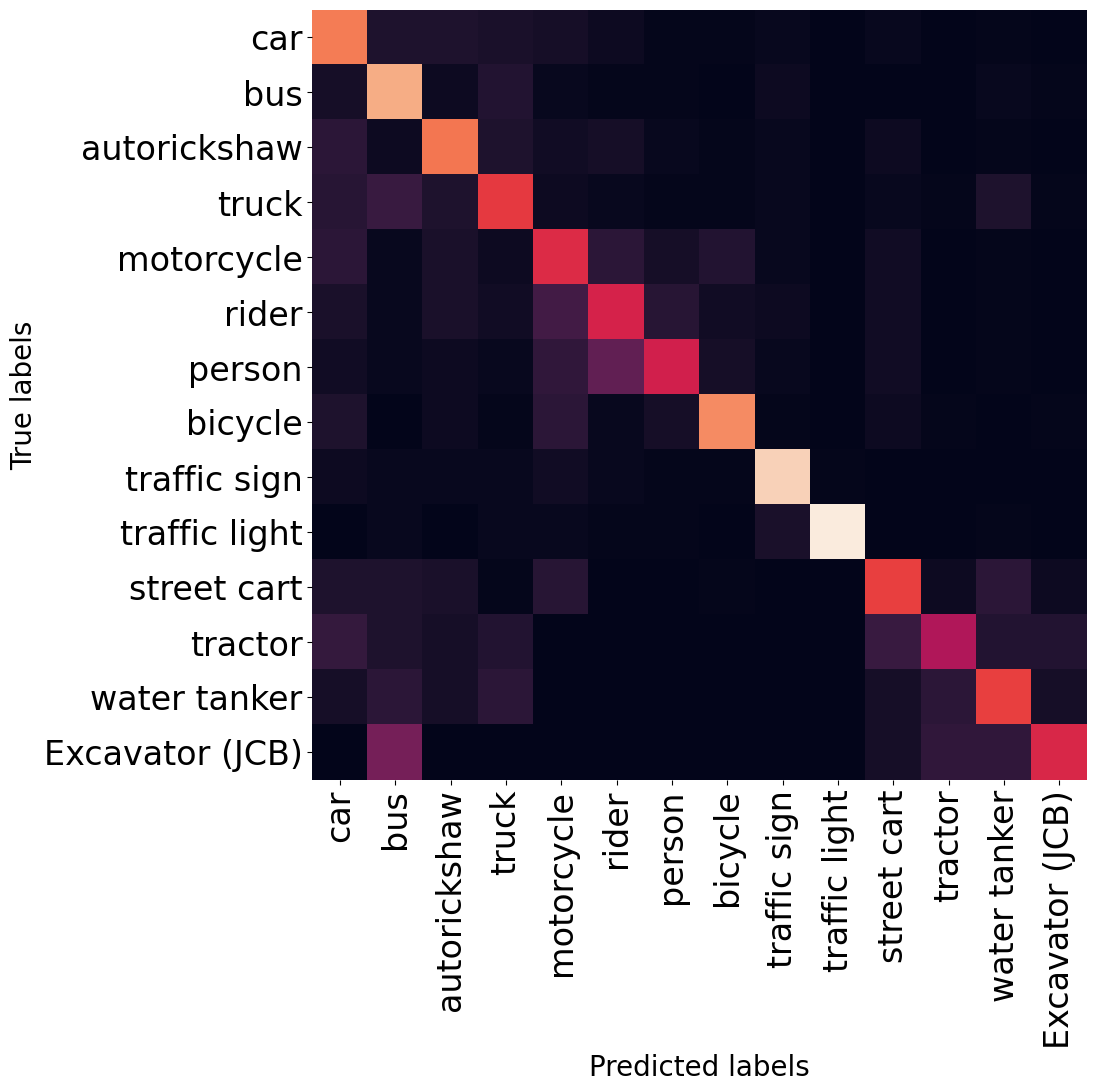} \\
                 (a) FsDet (Confusion = 68 \%) &
                 (b) FSCE (Confusion = 56 \%) &
                 (c) AGCM (Confusion = 46 \%)
                 
        \end{tabular}
        \caption{Confusion Matrix plot for the proposed AGCM technique. Our method shows a significant reduction (10\%) in class confusion between base and novel classes as compared to SoTA metric learning based FSOD techniques FSCE and FsDet.}
        \label{fig:cm}
\end{figure*}

\subsection{Components of the AGCM Architecture}
\label{abl:comp}
The AGCM approach consists of three main components. First, we adopt the stronger baseline of FSCE \cite{fscontrastive} which facilitates the inclusion of low-confidence proposals of the novel classes resulting in a significant performance gain over the best performing architecture demonstrated in \cite{majee2021fewshot} on IDD-OS. Secondly, our proposed APF module (refer section \ref{meth:frm}) reduces the effect of catastrophic forgetting by encouraging the formation of tighter class-specific feature clusters through attentive re-weighting of the RoI proposals. Finally, the cosine margin cross-entropy loss reduces the inter-class bias by increasing the angular margin between feature clusters. It ensures a reduction in confusion among object classes that share a large portion of low-level features. The quantitative contributions of each component is tabulated in Table \ref{tab:agcm_comp}.

\subsection{Ablation on key hyper-parameters in AGCM}
We perform ablation on various hyper-parameters introduced in our approach and derive their values which lead to the best possible novel class performance during the few-shot adaptation stage.


\textbf{Hyper-parameters of the APF Module :}
\label{sec:hyper_apf}
The hyper-parameter $\alpha$ introduced in section \ref{meth:frm} controls the ratio of the contribution of the chosen RoI with respect to other RoIs in the APF module. We vary the value for $\alpha$ between $\alpha = 0.5$ to $\alpha =  1.0$ and record the variation in performance of the novel classes in Table \ref{tab:hyper}. For smaller values of $\alpha$, there is a loss of distinctiveness for a feature proposal, and therefore, we see a loss in performance for both base and novel classes. On the other hand, for higher values of $\alpha$, no information propagation happens among the RoI proposals, which increases class confusion and deteriorates the performance on the base class. We thus chose $\alpha = 0.8$ for our experiments across all datasets.


Attentive weights ($w_{ij}$) computed for each RoI proposal through equation \ref{eq:attn} are calculated through a learnable metric. We ablate this metric in table \ref{tab:hyper} and record the variations in base and novel class performance while maintaining $\alpha$ at 0.8. We chose the cosine similarity metric over others as it achieves the best overall performance.

\textbf{Hyper-parameters of the Cosine Margin Cross-Entropy Loss :}
\label{sec:hyper_cos}
As shown in Table \ref{tab:hyper}, we vary the value of $m$ in the range of [0,1] and observe an increase in novel class performance between $m \geq 0$ to $m \leq 0.2$ followed by a decrease in both base and novel class performance from $m > 0.2$ to $m \leq 1.0$.  
Consequently, we adopt the value of $m$ as 0.2, which results in the highest overall performance gains for all our experiments across datasets. 
Although we observe a large gain in novel class performance (4 $mAP$ points), a small drop (1 $mAP$ point) is observed in the base class performance as the angular margin increases the inter-class intra-class bias among classes.

\subsection{Class Confusion Among Road Objects}
\label{abl:clsconf}
Figure \ref{fig:cm} shows the confusion matrix for our proposed AGCM in contrast with FSCE and FsDet approaches on all classes in IDD-OS for the 10-shot setting. FsDet shows a large confusion of 68.4\% while FSCE has 56\% confusion. AGCM achieves the least confusion of 46\%. Although FsDet can discriminate between the novel classes, it shows significant confusion among base classes with large intra-class variance like \emph{car}, \emph{bus} and \emph{truck}. The contrastive training strategy adopted by FSCE can reduce confusion between such classes but fails to overcome the inter-class bias between co-occurring classes like \emph{motorcycle}, \emph{person}, and \emph{rider}. AGCM overcomes the intra-class variance through proposal fusion (APF) while encouraging inter-class separation through margin penalties, reducing class confusion among classes. 



\begin{table}[h]
    \caption{\small Ablation experiment on catastrophic forgetting of base classes on IDD-OS split in the 10-shot setting. The $mAP_{base}$ before few-shot adaption is 63.4 $mAP$ points.}
    \centering
    \small
    \scalebox{0.9}{
    \begin{tabular}{|l|c|c|c|}
        \hline
         Method & $mAP_{base}$ & $mAP_{novel}$ & \% drop ($\downarrow$) \\
         \hline
         \small FRCNN-ft & & &\\
         (only base classes) & 63.4 & - & - \\
         \hline
         FsDet w/ cos & 47.8 & 39.8 & 24.6 \\
         FSCE  & 45.5 & 51.6 & 28.2 \\
         AGCM (ours) & \textbf{51.5} & \textbf{58.0} & \textbf{18.8}\\
         \hline
    \end{tabular}}
    \label{tab:abl_forget}
\end{table}

\subsection{Catastrophic Forgetting of Base Classes}
\label{abl:baseforget}
This section quantifies the drop in base class performance for multiple metric learning techniques and shows that our proposed AGCM approach achieves minimum degradation in base class performance while boosting the performance of novel classes. A Faster-RCNN model with a ResNet-101 backbone trained on 10 base classes in IDD-OS (referred as FRCNN-ft in table \ref{tab:abl_forget}) is used as the base model for all evaluated techniques. The results of this experiment after the few-shot adaptation stage (in 10-shot setting) are demonstrated through table \ref{tab:abl_forget}. Our method achieves the least degradation in base class performance of 18.8\% while obtaining the highest base and novel class performance.

\section{Conclusion}
In this work, we introduced a novel FSOD technique, Attention Guided Cosine Margin (AGCM), to overcome the class imbalance in Few-Shot Road Object Detection. Our method achieves State-of-The-Art (SoTA) results on all the splits of India Driving Dataset, outperforming the SoTA metric learners by up to 6.4 mAP points in IDD-OS split 10-shot setting. 
AGCM also generalizes to standard FSOD benchmarks like PASCAL-VOC, where we outperform SoTA approaches by up to 4.9 $mAP$ points. 
Our proposed Attention Proposal Fusion (APF) module minimizes catastrophic forgetting by 19\% by reducing intra-class variance. APF is computationally inexpensive and can be used with any two-stage detector. The introduced Cosine Margin Cross-Entropy loss increases the angular margin between overlapping classes reducing class confusion by 10\%. 

\bibliographystyle{ieee_fullname}
\bibliography{references}

\end{document}